\documentclass[runningheads]{llncs}
\usepackage{graphicx}
\usepackage{makecell}
\usepackage{booktabs}
\usepackage{tikz}
\usepackage{comment}
\usepackage{amsmath,amssymb} 
\usepackage{color}
\usepackage{orcidlink}

\begin{document}
\pagestyle{headings}
\mainmatter
\def\ECCVSubNumber{1578}  

\title{Contrastive Positive Mining for Unsupervised 3D Action Representation Learning} 

\titlerunning{CPM for Unsupervised 3D Action Representation Learning}
%
\author{Haoyuan Zhang (Corresponding author)\inst{1}\orcidlink{0000-0001-6895-3578}\index{Corresponding author} \and
	Yonghong Hou\inst{1}\orcidlink{0000-0002-1676-5505} \and
	Wenjing Zhang\inst{1}\orcidlink{0000-0002-9130-3829} \and
	Wanqing Li\inst{2}\orcidlink{0000-0002-4427-2687}}
\authorrunning{H. Zhang et al.}
%
\institute{Tianjin University, School of Electrical and Information Engineering, Tianjin, China\\ 
\email{\{zhy0860,houroy,zwj759\}@tju.edu.cn}\\ \and
Advanced Multimedia Research Lab, University of Wollongong, Wollongong, Australia.\\
\email{wanqing@uow.edu.au}}
\maketitle

\begin{abstract}
Recent contrastive based 3D action representation learning has made great progress. However, the strict positive/negative constraint is yet to be relaxed and the use of non-self positive is yet to be explored. In this paper, a Contrastive Positive Mining (CPM) framework is proposed for unsupervised skeleton 3D action representation learning. The CPM identifies non-self positives in a contextual queue to boost learning. Specifically, the siamese encoders are adopted and trained to match the similarity distributions of the augmented instances in reference to all instances in the contextual queue. By identifying the non-self positive instances in the queue, a positive-enhanced learning strategy is proposed to leverage the knowledge of mined positives to boost the robustness of the learned latent space against intra-class and inter-class diversity. Experimental results have shown that the proposed CPM is effective and outperforms the existing state-of-the-art unsupervised methods on the challenging NTU and PKU-MMD datasets.

\end{abstract}

\section{Introduction}

Human action recognition is an active research in recent years. Due to being light-weight, privacy-preserving and robust against complex conditions~\cite{shi2017learning,si2019attention,chen2021lstm,song2018spatio}, 3D skeleton is becoming a popular modality for capturing human action dynamics~\cite{hou2016skeleton,yan2018spatial,sun2021multi,zhang2020context}. Majority of previous skeleton-based action recognition approaches~\cite{liu20173d,wang2017scene,xiao2019action,zhang2020sar} are developed with a fully-supervised manner. However, in order to learn a good action representation, supervised methods require a huge number of labeled skeleton samples which is expensive and difficult to obtain. It impels the exploration of learning skeleton-based action representation in an unsupervised manner~\cite{lin2020ms2l,rao2021augmented,su2020predict,li20213d}. Often unsupervised methods use pretext tasks to generate the supervision signals, such as reconstruction~\cite{gui2018adversarial,zheng2018unsupervised}, auto-regression~\cite{kundu2019unsupervised,su2020predict} and jigsaw puzzles~\cite{noroozi2016unsupervised,wei2019iterative}. Consequently, the learning highly relies on the quality of the designed pretext tasks, and those tasks are hard to be generalized for different downstream tasks. Recent unsupervised methods employ advanced contrastive learning ~\cite{lin2020ms2l,rao2021augmented,li20213d} for instance discrimination in a latent space and have achieved promising results.

\begin{figure}
	\centering
	\includegraphics[height = 50mm]{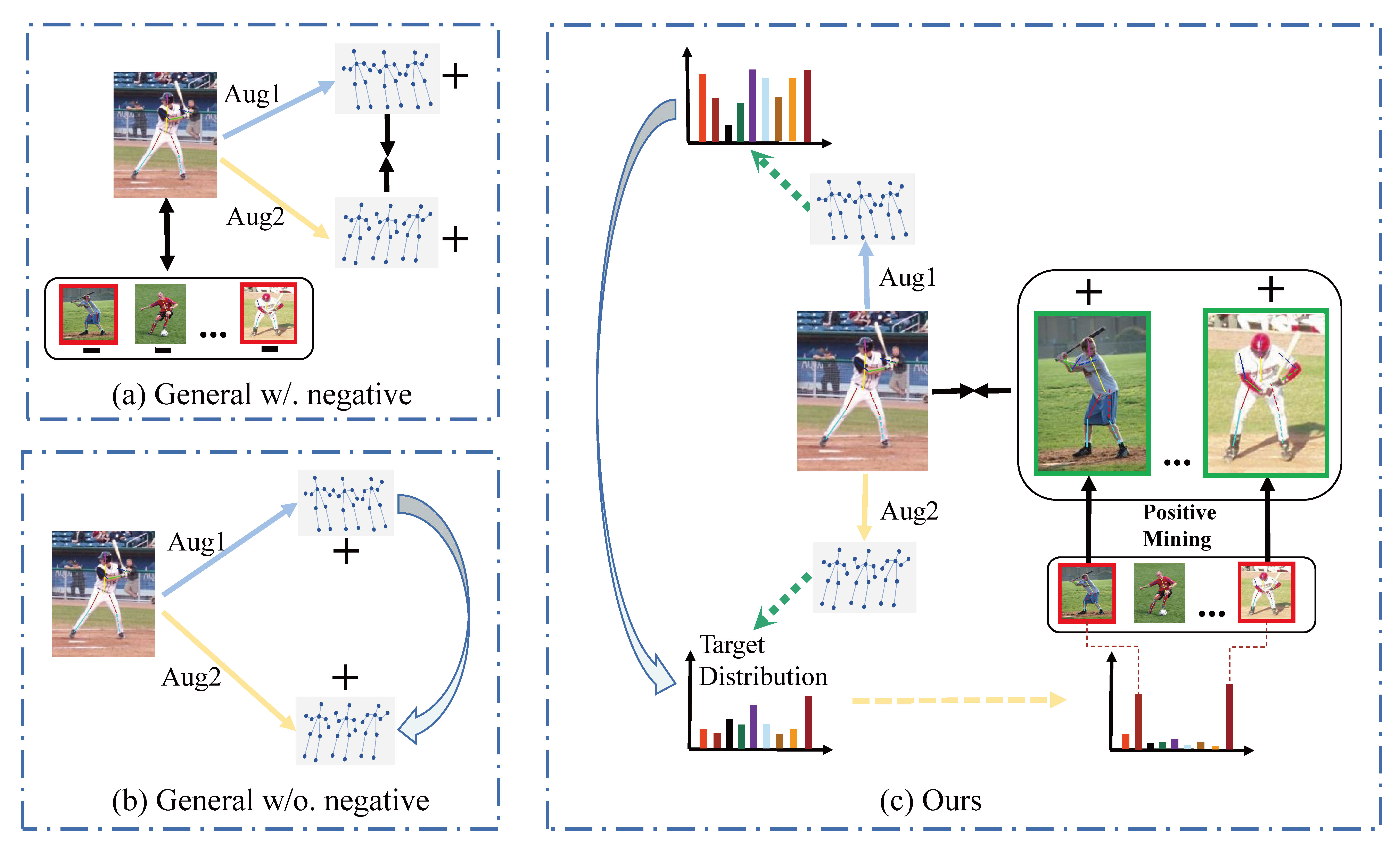}
	\caption{Illustrations about the proposed CPM and previous contrastive methods. (a) contrastive learning methods with negative~\cite{lin2020ms2l,rao2021augmented}. (b) contrastive learning methods without negative~\cite{chen2021exploring,grill2020bootstrap}. (c) the proposed Contrastive Positive Mining (CPM) method.}
	\label{fig:motivation}
\end{figure}

Although contrastive methods can improve the learning of skeleton representation, there are several issues, as illustrated in Fig.~\ref{fig:motivation}, in the current methods. Fig.~\ref{fig:motivation}(a) shows that the conventional contrastive learning methods require negatives~\cite{lin2020ms2l,rao2021augmented}. They only regard different augmentations of the same instance as positives to be drawn close during the learning, while other instances in the queue, usually formed by training samples in the previous round of epochs or batches, are all regarded as negatives and pushed apart from the current instance. Although these methods consider the correlation of current instance with others, there are inevitably instances in queue that belong to the same category as the current instance (marked with red rectangular box) and these instances are mistaken as negatives, which could degrade the learned representation. To address this issue, as shown in Fig.~\ref{fig:motivation}(c), this paper proposes to search for the instances in queue that are likely to be the same class of the current instance, then to consider those instances as non-self positives (marked with green rectangular box) and draw them close to current instance so as to improve the learning.

Fig.~\ref{fig:motivation}(b) shows the conventional contrastive learning methods without negatives~\cite{chen2021exploring,grill2020bootstrap}. The positive setting is similar to the previous methods illustrated in Fig.~\ref{fig:motivation}(a). Only different augmentations of individual instance are used as positive, consistency among current instance and the non-self instances with the same class are ignored during learning, limiting the representation ability for intra-class diversity. Besides, although non-negative manner avoids the instances of same class being pushed apart, the correlation of different instances are not considered. 

Notice that contrastive objective of both methods (i.e. with or without negatives) is on individual instances, which challenges learning a feature space for all instances. To overcome the above shortcomings, as illustrated in Fig.~\ref{fig:motivation}(c), the proposed method extends the contrastive objective from individual instances by keeping a queue of instances and mining the non-self positives in the queue to boost learning. Specifically, a novel Contrastive Positive Mining (CPM) framework is proposed for unsupervised skeleton 3D action recognition. The proposed CPM is a siamese structure with a student and a target branch, which follows the SimSiam~\cite{chen2021exploring}. The student network is trained to match the target network in terms of the similarity distribution of the augmented instance in reference to all instances in a contextual queue, so that the non-self positive instances with high similarity can be identified in the queue. Then a positive-enhanced learning strategy is proposed to leverage the mined non-self positives to guide the learning of the student network. This strategy boosts the robustness of the learned latent space against intra-class and inter-class diversity. Experimental results on NTU-60~\cite{shahroudy2016ntu}, NTU-120~\cite{liu2019ntu} and PKU-MMD~\cite{liu2020benchmark} datasets have validated the effectiveness of the proposed strategy.

To summarize, the key contributions include:
\begin{itemize}
	\item A novel Contrastive Positive Mining (CPM) framework for unsupervised learning of skeleton representation for 3D action recognition.
	\item A simple but effective non-self positive mining scheme to identify the positives in a contextual queue.
	\item A novel positive-enhanced leaning strategy to guide the learning of the student network via the target network.
	\item Extensive evaluation of the CPM on the widely used datasets, NTU and PKU-MMD, with state-of-the-art results obtained.
\end{itemize}

\section{Related Works}
\subsection{Unsupervised Contrastive Learning}
Contrastive learning is derived from noise-contrastive estimation~\cite{gutmann2012noise}, which contrasts different type of noises to estimate the latent distribution. It has been extended in different ways for unsupervised learning. Contrastive Prediction Coding (CPC)~\cite{oord2018representation} develops the info-NCE to learn image representation, with an auto-regressive model used to predict future in latent space. Contrastive Multiview Coding (CMC)~\cite{tian2020contrastive} leverages multi-view as positive samples, so that the information shared between multiple views can be captured by the learned representation. However, there often lacks of negative instances for the above methods. To solve this issue, a scheme called memory-bank~\cite{wu2018unsupervised} is developed in which the previous random representations are stored as negative instances, and each of them are regarded as an independent class. Recently, MoCo~\cite{he2020momentum} utilizes a dynamic dictionary to improve the memory-bank, and introduces the momentum updated encoder to boost the representation learning. Another way to enrich the negative instances is to use large batch-size such as in SimCLR~\cite{chen2020simple}. Particularly, SimCLR samples negatives from a large batch and shows that different augmentation, large batch size, and nonlinear projection head are all important for effective contrastive learning. However, these methods all regard different augmentations of the same instance as the only positives, while other instances in the queue including the ones with same category are all considered as negatives which cannot fully leverage capability of contrastive learning due to highly likely mixture of positives in the negatives. 

To deal with this issue, some negative-sample-free approaches are recently developed. SimSiam~\cite{chen2021exploring} shows that simple siamese twin networks with a stop-gradient operation to prevent collapsing can learn a meaningful representation. Barlow Twins~\cite{zbontar2021barlow} proposes an unsupervised objective function by measuring the cross-correlation matrix between the outputs of two identical networks. BYOL~\cite{grill2020bootstrap} learns a potentially enhanced representation from an online network by predicting the representation from a given representation learned from a target network with slow updating. However, these methods do not consider consistency learning among current instances and the non-self instances with the same class.

\subsection{Unsupervised 3D Action Recognition}
Unsupervised methods~\cite{srivastava2015unsupervised,luo2017unsupervised,li2018unsupervised} for video based action recognition are well developed, while few works are specifically for skeletons. LongT GAN (Generative adversarial network)~\cite{zheng2018unsupervised} is an auto-encoder-based GAN for skeleton sequence reconstruction. P\&C~\cite{su2020predict} employs an encoder-decoder learning structure, the encoder is weakened compared with decoder to learn more representative features. ASCAL~\cite{rao2021augmented} is a momentum LSTM with a dynamic updated memory-bank, augmented instances of the input skeleton sequence are contrasted to learn representation. MS$^2$L~\cite{lin2020ms2l} is a multi-task learning framework, with both pretext tasks and contrastive learning. CrosSCLR~\cite{li20213d} adopts a cross-view contrastive learning scheme and leverages multi-view complementary supervision signal. However, these methods either require pretext tasks or a large amount of negative samples, or rely on the reconstruction. 
\begin{figure}
	\centering
	\includegraphics[height = 45mm]{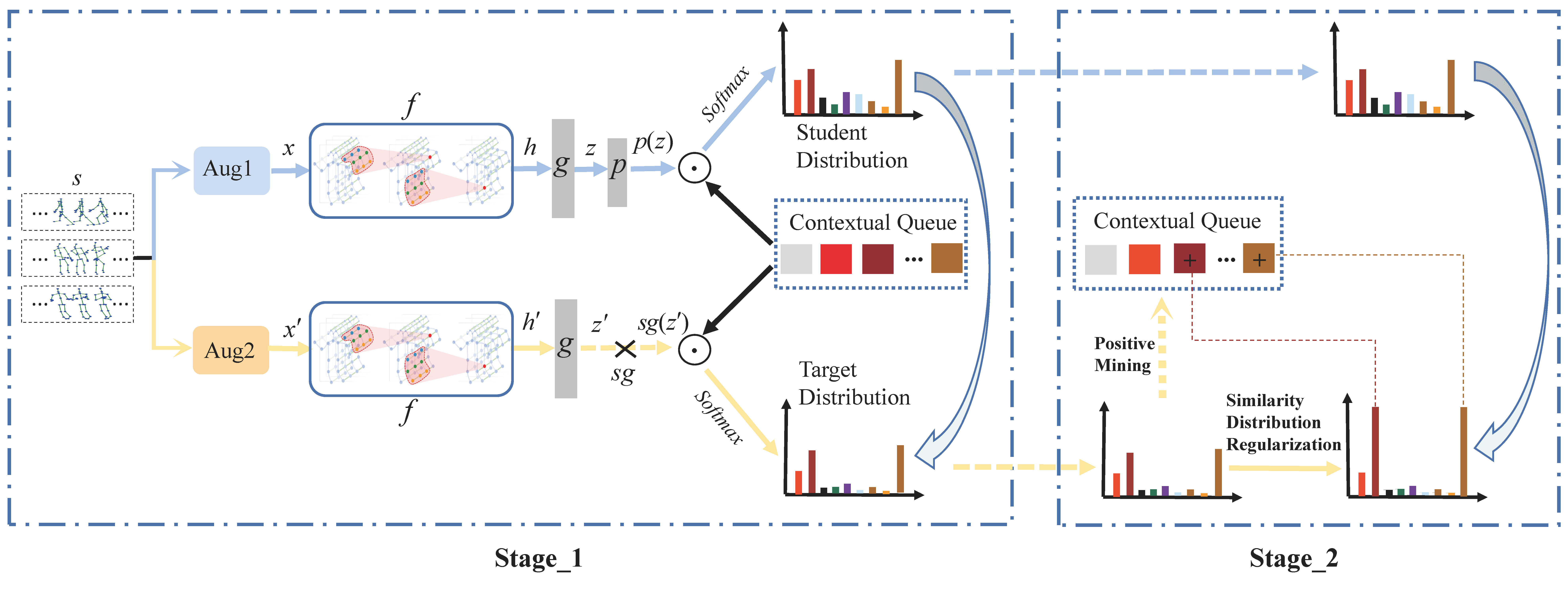}
	\caption{Overview of the CPM framework. CPM includes two stages. In the first stage, the student branch is trained to predict the inter-skeleton similarity distribution inferred by the target network, so as to excavate non-self positive. Then in the second stage the information of mined positives is injected into the target branch through similarity distribution regularization to guide the learning of student, which achieves positive-enhanced learning (different colors in distribution and contextual queue represent the embeddings of different instances, '+' means mined positive).}
	\label{fig:overview}
\end{figure}

\section{Proposed Method}
\label{sec:proposed_method}

\subsection{Overview}
Fig.~\ref{fig:overview} shows the basic framework of CPM. CPM adopts siamese twin networks as inspired by SimSiam~\cite{chen2021exploring}.
3D skeleton sequences are randomly augmented. Assume that a skeleton sequence $s$ has $T$ frames, $V$ joints, and $C$ coordinate channels, which can be represented as $s\in{{\rm{R}}^{C \times T \times V}}$. To augment $s$ into different versions $x$ and $x'$, a skeleton-specific augmentation strategy is needed. Different from the augmentations implemented for images, augmentation of skeleton sequences needs to be effective for learning spatial-temporal dynamics. In this paper, shear and crop in the spatial and temporal domain are to augment samples. Specifically, shear is applied as a spatial augmentation and is implemented as a linear transformation that displaces the skeleton joint in a fixed direction. Skeleton sequences are multiplied by a transformation matrix on the channel dimension, so as to slant the shape of 3D coordinates from body joints at a random angle. Crop is to pad a number of frames to a sequence symmetrically, then the sequence is randomly cropped into a fixed length~\cite{rao2021augmented,li20213d}.

The siamese encoders with identical network structure are to encode the augmented skeleton sequences, as shown in Fig.~\ref{fig:overview}, in a latent feature space. One branch is referred to as the student and the other serves as the target~\cite{tarvainen2017mean}. ST-GCN~\cite{yan2018spatial} is adopted as the encoder networks. 
The siamese encoders consist of several GCN layers and embed the two augmented skeleton sequences $x$ and $x'$ into a latent space. In each layer, human pose in spatial-dimension and joint's motion in temporal-dimension are alternatively encoded, i.e. a spatial graph convolution is followed by a temporal convolution.

After the siamese encoders, a projection MLP $g$ is attached to project the vector $h$ and $h'$ in the encoding space: $z = g(h)$, $z' = g(h')$, where $z$ and $z'$ are assumed to be mean-centered along the batch dimension so that each unit has 0 mean output over the batch. The projection MLP consists of two layers, the first one is followed by a batch normalization layer and rectified linear units. After the projection MLP, a prediction MLP $p$ with same architecture of $g$ is attached to the student branch to produce the prediction $p(z)$, while the stop-gradient operation is used in the target branch with the output $sg(z')$. In addition, a “first-in first-out”~\cite{he2020momentum} contextual queue $Q = [{a_1},...,{a_N}]$ is used to measure how well the encoded augmented instance by student network matches that by the target network with respect to the instances in the queue.

The key idea of the proposed method is to use the output of the student network to predict the output of the target network. More specifically, our objective is to train the siamese encoders such that the student network matches the target network in terms of the similarity distribution of the augmented instance in reference to all instances in the queue.

\subsection{Similarity Distribution and Positive Mining}

Similarity between the encoded feature of the augmentations and the instances in queue is first calculated and similarity distribution for the student network and the target network are calculated through softmax. The learning process is to train the network so that the similarity distribution of $x$ with respect to the instances in the queue can predict the distribution of $x'$. Compared with the previous methods, this strategy has the following advantages. No strict definition of positives/negatives is required and the match on similarity distribution over the instances in the queue is more reliable than that over individual instances. Since the similarity between the augmented instances and instances of the same class in the queue is expected to be high, resulting in an implicit mining of non-self positive instances in the queue.


Let $Q = [{a_1},...,{a_N}]$ be the queue of $N$ instances, where ${a_i}$ is the embedding of the $i$-th instance. The contextual queue comes from the preceding several batches of target network, which is updated in “first-in first-out”~\cite{he2020momentum} strategy.  Similarity distributions, $d_i$ and $d'_i$, between $\bar p(z)$ and ${a_i}$ and between  $\bar z'$ and ${a_i}$ are computed as follows, respectively,
\begin{align}
	{d_i} = \frac{{{e^{\bar p(z) \cdot {a_i}/\tau }}}}{{\sum\limits_{j = 1}^N {{e^{\bar p(z) \cdot {a_j}/\tau }}} }}\       \ \
	\label{eq1}
\end{align}

\begin{align}
	{d'_i} = \frac{{{e^{\bar z' \cdot {a_i}/\tau '}}}}{{\sum\limits_{j = 1}^N {{e^{\bar z' \cdot {a_j}/\tau '}}} }}      \ \
	\label{eq2}
\end{align}
where $\bar p(z)$ and $\bar z'$ are $l_2$ normalization of $p(z)$ and  $z'$. The overall similarity distributions, $D$ and $D'$ of the two arguments in the latent space with respect to the instances in the queue are, 
\begin{align}
	D = \left\{ {{d_i}} \right\}, D' = \left\{ {{d'_i}} \right\}, {\rm{ }}i \in N \ \
	\label{eq3}
\end{align}

The idea is to training siamese encoders to match $D$ with $D'$. In this paper, we adopt to minimize the Kullback-Leibler divergence between $D$ and $D'$, i.e.,

\begin{align}
	L = {D_{KL}}(D'||D) = H(D',D) - H(D') \ \
	\label{eq4}
\end{align}


By minimizing $L$, prediction $p(z)$ can be aligned with $z'$. Meanwhile, instances that belong to the same class could be pushed close in the latent space, while those from different classes are pushed apart. 


The similarity measures provide information for mining the positive instance in the queue. Specifically, given one instance's embedding $z$ and the corresponding queue $Q$, instances in queue with top-k high similarity are considered as positives, i.e.,

\begin{align}
	\Gamma \left( Q \right) = {\rm{Topk}}\left( Q \right) \ \
	\label{eq5}
\end{align}
which generates the index set of positive instances,
\begin{align}
	{D'_ + } = \left\{ {{d'_i}} \right\},{\rm{ }}i \in {N_ + } \ \
	\label{eq6}
\end{align}
where $N_ +$ is the index set of non-self positive instances. These positives can be used to facilitate a positive-enhanced learning as described below.

\subsection{Positive-enhanced Learning}
The non-self positives can be used to boost the representation learning. Intuitively, it is reasonable to inject the information of mined positives into the target branch to guide the learning of the student encoder. To do this, it is proposed to regularize the similarity distribution of the target branch ${D'}$ in each batch, so as to make use of the non-self positives iteratively. Specifically, we set the similarities of the $K$ mined positive instances in target branch to 1, which means those instances are considered the same action category with current instance. This strategy is referred to as ``positive-enhanced leaning''. The positive-enhanced similarity distribution can be expressed as, 
\begin{align}
	d^e_i = \left\{ {\begin{array}{*{20}{c}}
			{ \frac{{{e^{{1}/\tau' }}}}{{\sum\limits_{j = 1}^N {{e^{\bar z' \cdot {a_j}/\tau'}}} }}\ ,{\rm{ }}i \in {N_ + }}\\
			{d'_i,{\rm{ otherwise}}}
		\end{array}} \right. \ \
		\label{eq7}
\end{align}
	
Then we train distribution of student to continue predicting the regularized target distribution, so that the student is guided to learn more informative intra-class diversity brought by the non-self skeleton positives knowledge we inject,
\begin{align}
		L' = H({D'_{NP}},D) - H({D'_{NP}}) \ \
		\label{eq8}
\end{align}
where $D'_{NP}=\{d^e_i\}$ is the non-self positive-enhanced target distribution. Compared to Eq.(4), Eq.(8) intends to pull positive instances closer. 
	
\subsection{Learning of CPM}
In the early training stage, the model is likely not stable and capable enough of providing reasonable measures of the similarity distribution to identify the positives in the queue. Therefore, a two-stage training strategy is adopted for CPM: the student branch is first trained to predict the similarity distribution inferred by the target network without enhanced positives in Eq.(4). When it is stable, the model is trained using the positive-enhanced learning strategy in Eq.(8). 

\section{Experiments}
\label{sec:results}
\subsection{Datasets}

\textbf{NTU RGB+D 60 (NTU-60) Dataset~\cite{shahroudy2016ntu}}: NTU-60 is one of the widely used indoor-captured datasets for human action recognition. 56880 action clips in total are performed by 40 different actors in 60 action categories. The clips are captured by three cameras simultaneously at different horizontal angles and heights in a lab environment. Experiments are conducted on the Cross-Subject (X-Sub) and Cross-View (X-View) benchmarks. 

\textbf{NTU RGB+D 120 (NTU-120) Dataset~\cite{liu2019ntu}}: NTU-120 is an extended version of NTU-60. There are totally 114480 action clips in 120 action categories. Most settings of NTU-120 follow the NTU-60. Experiments are conducted on the Cross-Subject (X-Sub) and Cross-Setup (X-Set) benchmarks.  

\textbf{PKU-MMD Dataset~\cite{liu2020benchmark}}: There are nearly 20,000 action clips in 51 action categories. Two subsets PKU-MMD I and PKU-MMD II are used in the experiments. PKU-MMD II is more challenging than PKU-MMD I as it has higher level of noise. Experiments are conducted on the Cross-Subject (X-Sub) benchmark for both subsets.

\subsection{Implementation}

\textbf{Architecture}: The 9-layer ST-GCN~\cite{yan2018spatial} network is chosen as the encoders. In each layer, the spatial graph convolution is followed by a temporal convolution, the temporal convolutional kernel size is 9. A projector of 2-layer MLP is attached to the output of both networks. The first layer is followed by a batch normalization layer and rectified linear units, with output size of 512, while the output dimension of the second layer is 128. A predictor with the same architecture is used in the student branch, while the stop-gradient operation is applied in target branch. The contextual queue size $N$ is set to $65536$, $32768$ and $16384$ for NTU-60/120, PKU-MMD I and PKU-MMD II datasets, respectively.

\textbf{Unsupervised Pre-training}: LARS~\cite{you2017large} is utilized as optimizer and trained for 400 epochs with batch size 512, note that the positive-enhanced learning is conducted after 300 epochs. The learning rate starts at 0 and is linearly increased to 0.5 in the first 10 epochs of training and then decreased to 0.0005 by a cosine decay schedule~\cite{loshchilov2016sgdr}. All experiments are conducted on one Nvidia RTX3090 GPU using PyTorch. 

\textbf{Linear Evaluation Protocol}: The pre-trained models are verified by linear evaluation. Specifically, a linear classifier (a fully-connected layer followed by a softmax layer) is trained supervisedly for 100 epochs while the pre-trained model is fixed.

\subsection{Results and Comparison}

\textbf{Unsupervised Results}: The performance of the proposed CPM is compared with the state-of-the-art supervised and unsupervised methods on the NTU and PKU-MMD datasets and results are shown in Table~\ref{tab1}. Following the standard practice in literature the recognition performance in terms of top-1 classification accuracy is reported. Note that, if not specified, the experiments including ablation study are conducted on the joint data. 3S means the ensemble results of joint, bone and motion data. 
The obvious performance improvement compared with the recent advanced unsupervised counterparts~\cite{li20213d,thoker2021skeleton} has been obtained and demonstrates the effectiveness of CPM. In addition, CPM (3S) outperforms the supervised ST-GCN~\cite{yan2018spatial} on both NTU and PKU-MMD datasets.

\setlength{\tabcolsep}{4pt}
\begin{table}
\begin{center}
\caption{Performance and comparison with the state-of-the-art methods on the NTU and PKU-MMD datasets.}
\label{tab1}
		\begin{tabular}{lllllll}
			\hline
			\multicolumn{1}{c}{Architectures}
			&\multicolumn{2}{c}{NTU-60 (\%)}
			&\multicolumn{2}{c}{NTU-120 (\%)} 
			&\multicolumn{2}{c}{PKU-MMD (\%)} \\
			\cmidrule(l){2-3} 	\cmidrule(l){4-5} \cmidrule(l){6-7}
			\multicolumn{1}{c}{}
			&\multicolumn{1}{c}{X-Sub} & {X-View}
			&\multicolumn{1}{c}{X-Sub} & {X-Set} 
			&\multicolumn{1}{c}{Part I} & {Part II} \\
			\hline
			\slshape{Supervised}&&&&&&  \\
			C-CNN + MTLN~\cite{ke2017new}&\makecell[c]{79.6}&\makecell[c]{84.8}&\makecell[c]{-}&\makecell[c]{-}&\makecell[c]{-} &\makecell[c]{-}  \\
			TSRJI~\cite{caetano2019skeleton}&\makecell[c]{73.3}&\makecell[c]{80.3}&\makecell[c]{67.9}&\makecell[c]{62.8}&\makecell[c]{-} &\makecell[c]{-} \\			
			ST-GCN~\cite{yan2018spatial}&\makecell[c]{81.5} &\makecell[c]{88.3}&\makecell[c]{70.7} &\makecell[c]{73.2}&\makecell[c]{84.1} &\makecell[c]{48.2} \\
			\hline
			\slshape{Unsupervised}&&&&&&  \\
			LongT GAN~\cite{zheng2018unsupervised}&\makecell[c]{39.1}&\makecell[c]{48.1}&\makecell[c]{-}&\makecell[c]{-}&\makecell[c]{67.7}&\makecell[c]{27.0} \\
			ASCAL~\cite{rao2021augmented}&\makecell[c]{58.5}&\makecell[c]{64.8}&\makecell[c]{48.6}&\makecell[c]{49.2}&\makecell[c]{-} &\makecell[c]{-} \\
			MS$^2$L~\cite{lin2020ms2l}&\makecell[c]{52.6}&\makecell[c]{-}&\makecell[c]{-}&\makecell[c]{-}&\makecell[c]{64.9} &\makecell[c]{27.6}\\
			P\&C~\cite{su2020predict}&\makecell[c]{50.7}&\makecell[c]{76.3}&\makecell[c]{-}&\makecell[c]{-}&\makecell[c]{-} &\makecell[c]{-}\\
			ISC~\cite{thoker2021skeleton}&\makecell[c]{76.3}&\makecell[c]{85.2}&\makecell[c]{67.9}&\makecell[c]{67.1}&\makecell[c]{80.9}&\makecell[c]{36.0}\\
			CrosSCLR (joint)~\cite{li20213d}&\makecell[c]{72.9}&\makecell[c]{79.9}&\makecell[c]{-}&\makecell[c]{-}&- &\makecell[c]{-}\\
			CrosSCLR (3S)~\cite{li20213d}&\makecell[c]{77.8}&\makecell[c]{83.4}&\makecell[c]{67.9} &\makecell[c]{66.7}&\makecell[c]{84.9}&\makecell[c]{-}\\
			CPM (joint)&\makecell[c]{78.7}&\makecell[c]{84.9}&\makecell[c]{68.7}&\makecell[c]{69.6}&\makecell[c]{88.8}&\makecell[c]{48.3}\\
			CPM (3S)&\makecell[c]{\textbf{83.2}} &\makecell[c]{\textbf{87.0}}&\makecell[c]{\textbf{73.0}} &\makecell[c]{\textbf{74.0}}&\makecell[c]{\textbf{90.7}}&\makecell[c]{\textbf{51.5}}\\
			\hline
		\end{tabular}
\end{center}
\end{table}	
\setlength{\tabcolsep}{1.4pt}
	
\textbf{Semi-supervised Results}: The CPM is first pre-trained on all training data in an unsupervised way, then the classifier is fine-tuned with 1\% and 10\% annotated data respectively. Table~\ref{tab2} shows the semi-supervised results on the NTU-60 dataset. The results have shown the proposed CPM performs significantly better than the compared methods. 
Compared with MS$^2$L~\cite{lin2020ms2l} and ISC~\cite{thoker2021skeleton}, CPM improves the performance by a large margin and shows its robustness when fewer labels are available for fine-tuning.	
\setlength{\tabcolsep}{4pt}	
\begin{table}
	\centering
	\caption{Semi-supervised performance and comparison with the state-of-the-art methods on the NTU-60 dataset.}\label{tab2}
	\begin{tabular}{llll}
		\hline
		Architectures&Label fraction (\%)&X-Sub (\%)&X-View (\%)\\
		\hline
		LongT GAN~\cite{zheng2018unsupervised}&\makecell[c]{1}&\makecell[c]{35.2} &\makecell[c]{-}\\
		MS$^2$L~\cite{lin2020ms2l}&\makecell[c]{1}&\makecell[c]{33.1}&\makecell[c]{-}\\
		ISC~\cite{thoker2021skeleton}&\makecell[c]{1}&\makecell[c]{35.7}&\makecell[c]{38.1}\\
		CPM&\makecell[c]{1}&\makecell[c]{\textbf{56.7}} &\makecell[c]{\textbf{57.5}}\\
		\hline
		LongT GAN~\cite{zheng2018unsupervised}&\makecell[c]{10}&\makecell[c]{62.0} &\makecell[c]{-}\\
		MS$^2$L~\cite{lin2020ms2l}&\makecell[c]{10}&\makecell[c]{65.2} &\makecell[c]{-}\\
		ISC~\cite{thoker2021skeleton}&\makecell[c]{10}&\makecell[c]{65.9}&\makecell[c]{72.5}\\
		CPM&\makecell[c]{10}&\makecell[c]{\textbf{73.0}} &\makecell[c]{\textbf{77.1}}\\
		\hline
	\end{tabular}
\end{table}	
\setlength{\tabcolsep}{1.4pt}	
		
\textbf{Fully Fine-tuned Results}: The model is first unsupervisedly pre-trained, then a linear classifier is appended to the learnable encoder. Both the pre-trained model and the classifier undergo a supervised training using all training data~\cite{zbontar2021barlow}, results are shown in Table~\ref{tab3}. 
On both NTU-60 and NTU-120 datasets the fully fine-tuned CPM outperforms the supervised ST-GCN~\cite{yan2018spatial}, demonstrating the effectiveness of the unsupervised pretraining.
\setlength{\tabcolsep}{4pt}
\begin{table}
	\centering
	\caption{Fully fine-tuned performance and comparison on the NTU-60 and NTU-120 datasets.}\label{tab3}
	\begin{tabular}{lllll}
		\hline
		\multicolumn{1}{c}{Architectures}
		&\multicolumn{2}{c}{NTU-60 (\%)}
		&\multicolumn{2}{c}{NTU-120 (\%)} \\
		\cmidrule(l){2-3} 	\cmidrule(l){4-5}
		\multicolumn{1}{c}{}
		&\multicolumn{1}{c}{X-Sub} & {X-View}
		&\multicolumn{1}{c}{X-Sub} & {X-Set} \\
		\hline
		C-CNN + MTLN~\cite{ke2017new}&\makecell[c]{79.6} &\makecell[c]{84.8}&\makecell[c]{-}&\makecell[c]{-} \\
		TSRJI~\cite{caetano2019skeleton}&\makecell[c]{73.3}&\makecell[c]{80.3}&\makecell[c]{67.9}&\makecell[c]{62.8} \\
		ST-GCN~\cite{yan2018spatial}&\makecell[c]{81.5} &\makecell[c]{88.3}&\makecell[c]{70.7}&\makecell[c]{73.2} \\
		CPM&\makecell[c]{\textbf{84.8}}&\makecell[c]{\textbf{91.1}}&\makecell[c]{\textbf{78.4}}&\makecell[c]{\textbf{78.9}}\\
		\hline
	\end{tabular}
\end{table}
\setlength{\tabcolsep}{1.4pt}

\subsection{Ablation Study}\	
\textbf{On positive mining}: To verify the effectiveness of positive-enhanced learning, we pre-train the CPM (w/o. PM) without identifying the positive instances and, hence, positive-enhanced learning, other settings are kept the same. Performance of CPM and CPM (w/o. PM) is shown in Table~\ref{tab4}. On the NTU-60 X-Sub and X-View tasks, CPM improves the recognition accuracy by $3.1$ percentage points and $3.2$ percentage points, respectively. On the NTU-120 X-Sub and X-Set tasks, $3.9$ percentage points and $4.9$ percentage points improvements are obtained by CPM. This demonstrates that identification of positive instances and the positive-enhanced learning strategy do improve the representation learning.

		\setlength{\tabcolsep}{4pt}			
		\begin{table}
				\centering
				\caption{Benefit of positive mining.}\label{tab4}
		\begin{tabular}{lll}
				\hline
				Datasets& CPM (w/o. PM) (\%)& CPM (\%)\\
				\hline 
				X-Sub (NTU-60)&\makecell[c]{75.6}&\makecell[c]{78.7}\\			
				X-View (NTU-60)&\makecell[c]{81.7}&\makecell[c]{84.9}\\
				X-Sub (NTU-120)&\makecell[c]{64.8}&\makecell[c]{68.7}\\
				X-Set (NTU-120)&\makecell[c]{64.7}&\makecell[c]{69.6}\\
				\hline
				\end{tabular}	
		\end{table}
		\setlength{\tabcolsep}{1.4pt}

\begin{figure}
	\centering
	\includegraphics[height = 50mm]{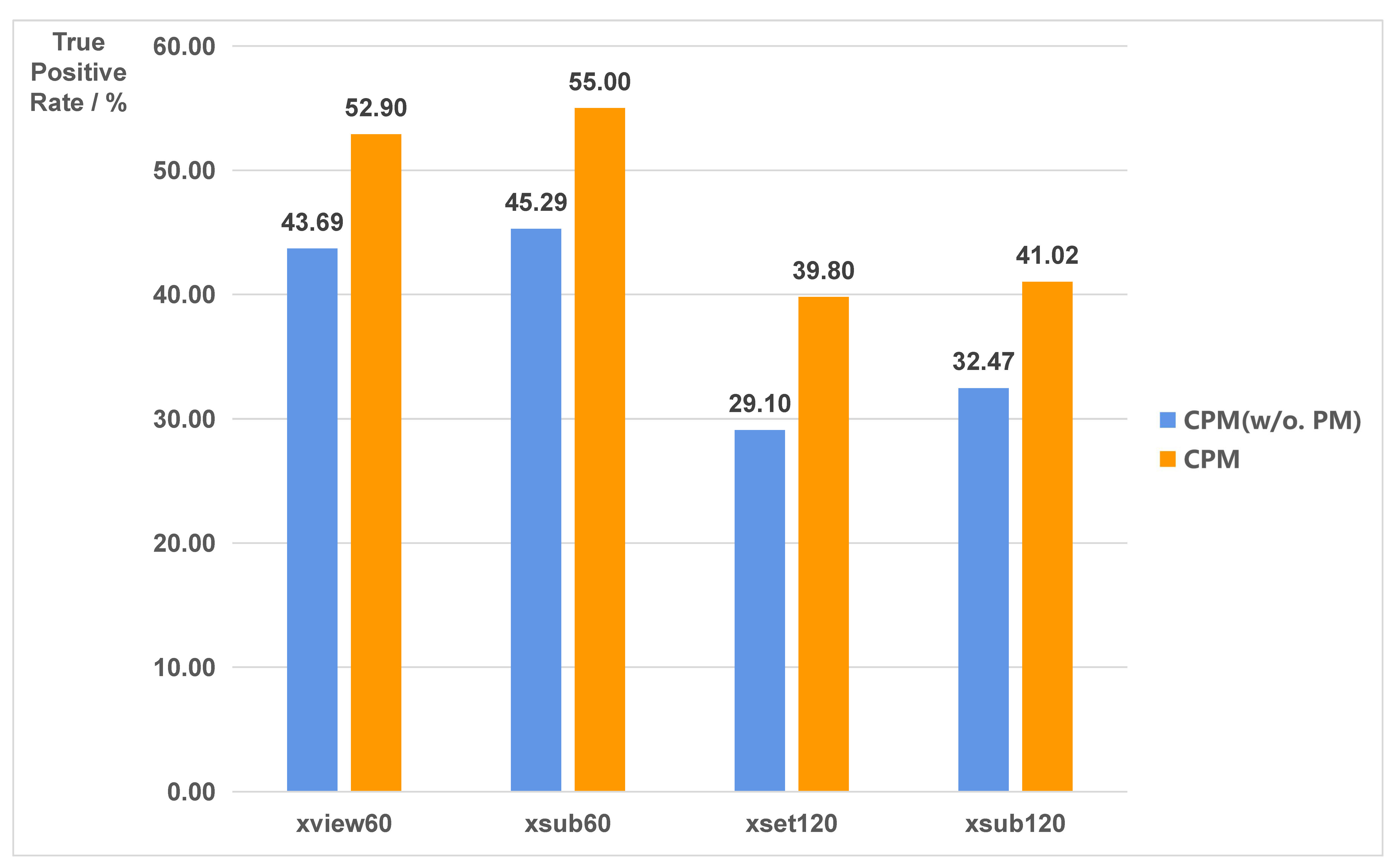}
	\caption{Precision of positive instances identified by CPM and CPM (w/o. PM) on different datasets.}
	\label{fig:Statistical Analysis}
\end{figure}

To further verify how well the non-self positives in the queue can be identified for the positive-enhanced learning, Fig.~\ref{fig:Statistical Analysis} shows the precision of the positives selected by CPM and CPM (w/o. PM) in one epoch in the top-100 identified positive instances. The results show that even CPM (w/o. PM) is capable of identifying many true positives. This is in significant contrast to the methods in~\cite{lin2020ms2l,rao2021augmented} where all instances in the queue would be considered as negatives. When positive-enhanced learning is applied, the precision has been significantly increased and so that the learned representation is more robust against the intra-class diversity. 
\begin{figure}
	\centering
	\includegraphics[height = 45mm]{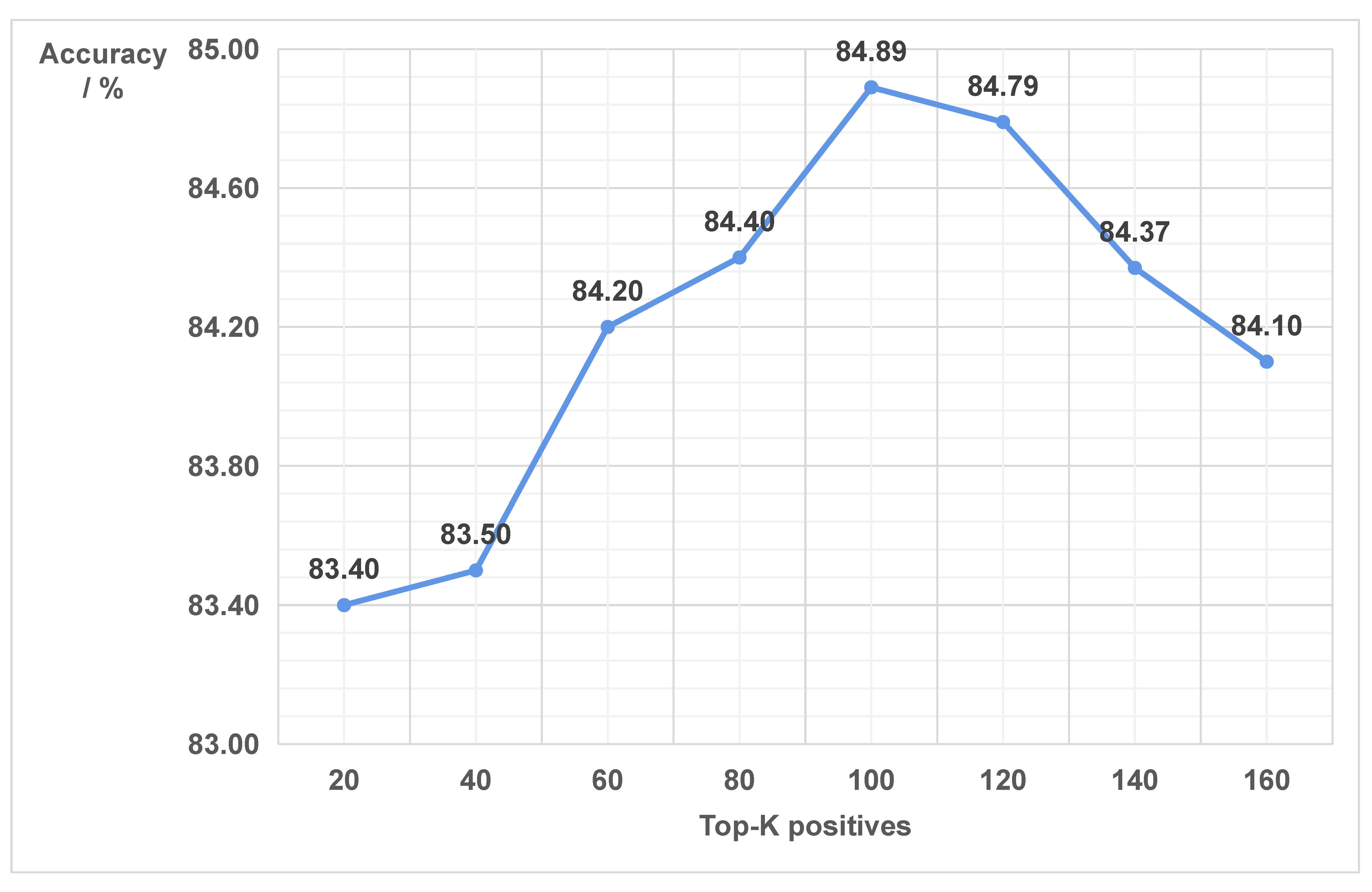}
	\caption{Effect of the number $K$ of top positives on the proposed CPM in the NTU-60 X-View task.}
	\label{fig:top-K mined positive}
\end{figure}	
				
\textbf{On the value of $K$}: Hyper-parameter $K$ refers to the number of positives identified in the queue. This study shows how $K$ affects the performance. Experiments have shown that when $K$ is 100, best results are obtained on the NTU-60 and NTU-120 datasets. Results on NTU-60 X-View are shown in Fig.~\ref{fig:top-K mined positive}. It is found that too large or too small $K$ both decreases the performance. A large value of $K$ could include unexpected false positives with low similarity that misleads the learning. A small value of $K$ might ignore too many true positives that would potentially decrease representation ability to accommodate intra-class diversity. Good performance was observed when $K$ is $50$ and $25$ for PKU-MMD I and PKU-MMD II datasets, respectively. It is conjectured that choice of $K$ may depend on the scale of the dataset. 

\begin{figure}
	\centering
	\includegraphics[height = 45mm]{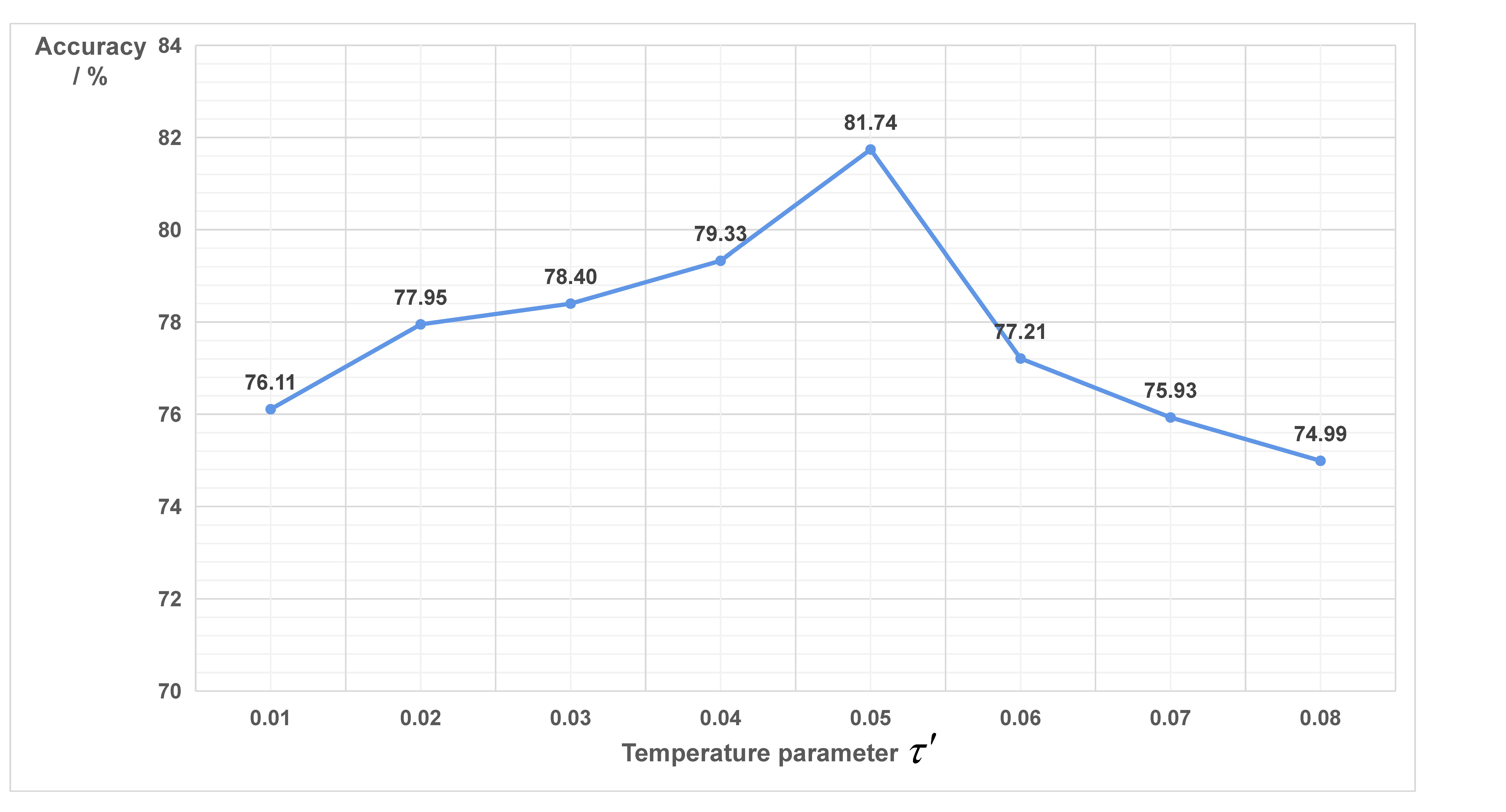}
	\caption{Effect of different temperature $\tau'$ on the performance in the NTU-60 X-View task.}
	\label{fig:temperature parameter}
\end{figure}											
\textbf{On the value of $\tau'$}: Fig.~\ref{fig:temperature parameter} shows the performance of CPM (w/o. PM) using different $\tau'$ with $\tau$ fixed to 0.1~\cite{fang2021seed}, the optimal performance is obtained when $\tau'$ is 0.05. A large value of $\tau'$ could lead to a flatter target distribution so that the learned representation becomes less discriminative. A small value of $\tau'$ would suppress the difference in similarities between the positive and the negative, leading to many false positives included in the positive-enhanced learning. If $\tau'$ is too small, less positive instances could be identified and this would again adversely affect the effectiveness of learning. 

\begin{figure}
	\centering
	\includegraphics[height = 40mm]{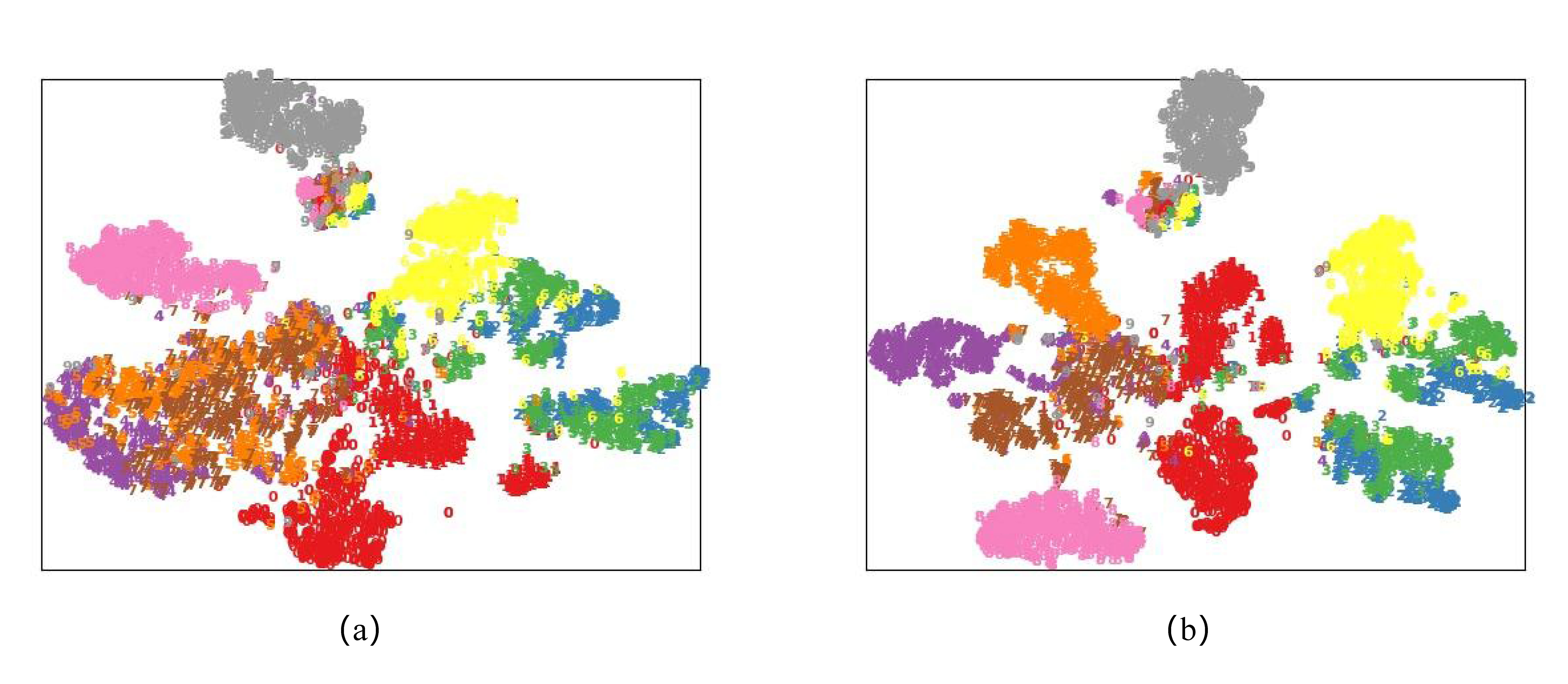}
	\caption{t-SNE visualization of embedding for (a) CPM (w/o. PM) and (b) CPM on the NTU-60 X-View task (best view in color).}
	\label{fig:tsne}
\end{figure}	
								
\textbf{Embedding Visualization}: t-SNE~\cite{van2008visualizing} is used to visualize the embedding clustering produced by CPM (w/o. PM) and CPM as shown in Fig.~\ref{fig:tsne}. Note that embedding of $10$ different action categories are sampled and visualized with different colors. The visual results show how well the embedding of the same type of actions form clusters while different types of actions are separated. By comparing the t-SNE of CPM and CPM (w/o. PM), CPM has clearly improved the clustering of actions, which indicates that the learned latent space is more discriminative than the space learned without using positive-enhanced learning strategy. 
																		
\section{Conclusion}
\label{sec:conclusion_future_work}
In this paper, a novel unsupervised learning framework called Contrastive Positive Mining (CPM) is developed for learning 3D skeleton action representation. The proposed CPM follows the SimSiam~\cite{chen2021exploring} structure, consisting of siamese encoders, student and target. By constructing a contextual queue and identifying non-self positive instances in the queue, the student encoder is able to learn a discriminative latent space by matching the similarity distributions of individual instance's two augments with respect to the instances in the queue. In addition, by identifying positive instances in the queue, a positive-enhanced learning strategy is developed to boost the robustness of the learned latent space against intra-class and inter-class diversity. Experiments on the NTU and PKU-MMD datasets have shown that the proposed CPM obtains the state-of-the-art results.

\clearpage
%
%
\bibliographystyle{splncs04}
\bibliography{egbib}
\end{document}